%% file: templateArxiv.tex
\documentclass{article}
\usepackage{graphicx}  
\usepackage{colortbl} 
\usepackage[table]{xcolor} 
\usepackage{array}    
\usepackage{amsmath}  
\usepackage{multirow} 
\definecolor{lightgray}{HTML}{EFEFEF}
\definecolor{green}{HTML}{32CB00}
\definecolor{red}{HTML}{CB0000}
\usepackage{PRIMEarxiv}
\usepackage{amsmath}
\usepackage[utf8]{inputenc} 
\usepackage[T1]{fontenc}    
\usepackage{hyperref}       
\usepackage{url}            
\usepackage{booktabs}       
\usepackage{amsfonts}       
\usepackage{nicefrac}       
\usepackage{microtype}      
\usepackage{lipsum}
\usepackage{fancyhdr}       
\usepackage{graphicx}       
\graphicspath{{media/}}     

\title{Beyond Task Vectors: Selective Task Arithmetic Based on Importance Metrics}

\author{
Bowen Tian\thanks{The first two authors contributed equally to this work.}\\
HKUST(GZ)\thanks{Hong Kong University of Science and Technology(Guangzhou)}\\
\textit{$DI^2 Lab$}\thanks{Deep Interdisciplinary Intelligence Lab.}\\
{\tt\small bowentian@hkust-gz.edu.cn}
\and
Songning Lai\footnotemark[1]\\
HKUST(GZ)\\
\textit{$DI^2 Lab$}\\
{\tt\small songninglai@hkust-gz.edu.cn}
\and
Jiemin Wu\\
HKUST(GZ)\\
\textit{$DI^2 Lab$}\\
{\tt\small jieminwu@hkust-gz.edu.cn}
\and
Zhihao Shuai\\
HKUST(GZ)\\
{\tt\small zhihaoshuai@hkust-gz.edu.cn}
\and
Shiming Ge\\
Institute of Information Engineering\\
Chinese Academy of Sciences\\
{\tt\small geshiming@iie.ac.cn}
\and
Yutao Yue\thanks{Correspondence to Yutao Yue \{yutaoyue@hkust-gz.edu.cn\}}\\
HKUST(GZ)\\
\textit{$DI^2 Lab$}\\
\textit{Institute of Deep Perception Technology, JITRI}\\
{\tt\small yutaoyue@hkust-gz.edu.cn}\\
\thanks{This work was supported by Guangzhou-HKUST(GZ) Joint Funding Program(Grant No.2023A03J0008), Education Bureau of Guangzhou Municipality}
}

\begin{document}
\maketitle

\input{TaskVectorX/sec/0_abstract}

\input{TaskVectorX/sec/intro-new}
\input{TaskVectorX/sec/3_prelim}

\input{TaskVectorX/sec/3_method}

\input{TaskVectorX/sec/4_exp}

\input{TaskVectorX/sec/5_con}

\bibliographystyle{unsrt}

\end{document}

%% file: TaskVectorX/sec/0_abstract.tex
\begin{abstract}
Pretrained models have revolutionized deep learning by enabling significant performance improvements across a wide range of tasks, leveraging large-scale, pre-learned knowledge representations. However, deploying these models in real-world multi-task learning (MTL) scenarios poses substantial challenges, primarily due to high computational costs and inefficiencies in inference. Traditional approaches such as pruning, quantization, and knowledge distillation have been explored to mitigate these issues, but they often fall short in fully addressing the complexities of multi-task environments. This paper introduces \textbf{\underline{S}}elective \textbf{\underline{T}}ask \textbf{\underline{A}}rithmetic \underline{\textbf{(STA)}}, a training-free framework designed to enhance multi-task performance through task-specific parameter fusion. STA addresses three key challenges: (i) \textbf{Parameter importance diversity: } Recognizing that different tasks relie on distinct parameters, STA employs a loss-sensitive parameter importance metric derived from a first-order Taylor expansion to accurately measure the importance of parameters for each task. (ii) \textbf{Over-reliance on hyperparameter tuning: }By enhancing the sparsity of task vectors through parameter importance metrics, STA reduces the need for extensive hyperparameter tuning, thereby improving the generalization and robustness of the model. (iii) \textbf{Neglect of other abilities in task arithmetic: } Previous works have largely overlooked the potential for more precise task forgetting. STA leverages its parameter importance metric to achieve more controlled and effective task forgetting, minimizing the impact of noisy elements that can degrade model performance. Experimental results demonstrate that STA achieves superior multi-task performance across benchmarks and excellent performance in task forgetting.
\end{abstract}

%% file: TaskVectorX/sec/intro-new.tex
\section{Introduction}
\label{sec:intro}

In recent years, pretrained models have become foundational in deep learning, significantly enhancing performance across a range of tasks due to their ability to leverage large-scale, pre-learned knowledge representations. Despite this success, deploying these models in real-world multi-task learning (MTL) scenarios presents substantial challenges, particularly regarding computational cost and inference efficiency. These issues have spurred extensive research into model compression and optimization techniques, including pruning, quantization, and knowledge distillation. Among these, parameter-level task arithmetic \cite{ilharco2022editing} has emerged as a promising approach. This method allows models to encapsulate essential knowledge without additional training, thereby reducing deployment costs by minimizing parameter size while maintaining performance. However, effectively integrating task-specific parameters in a multi-task context remains challenging due to the inevitable noise introduced during the fusion process (e.g., when parameters of different symbols are fused, the result may not have a clear meaning).

\begin{figure}[t]
  \centering
  \includegraphics[width=0.7\columnwidth]{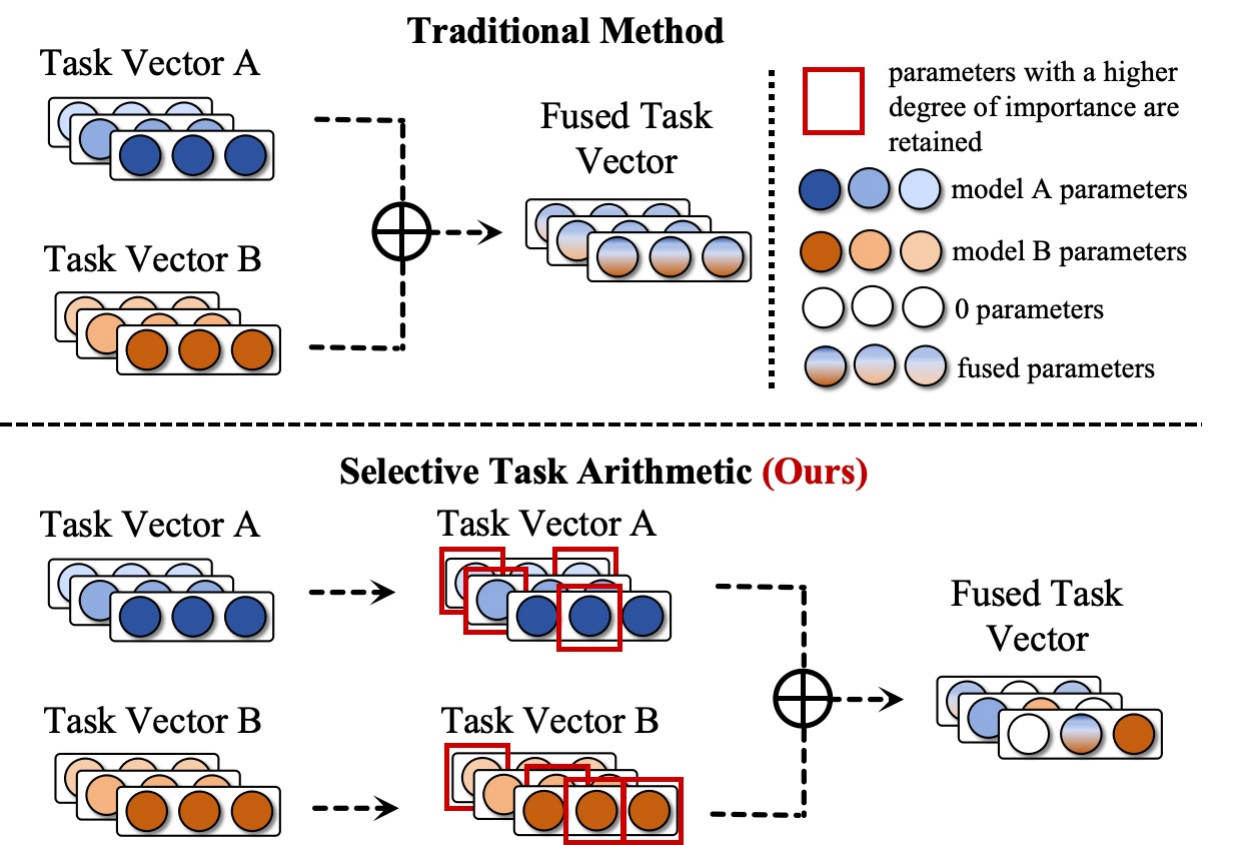}
  \caption{Our STA method allows for more precise control of task vectors for better task arithmetic.}
  \label{fig:intro}
\end{figure}

Given shared pretrained weights for different downstream models, the task arithmetic method calculates the \textbf{task vector} for each task by computing the difference between the fine-tuned weights and the pretrained weights. This method multiplies the sum of these task vectors by a coefficient and adds them to the pretrained weights to construct a model capable of performing multiple downstream tasks. Recently, several studies have proposed methods to enhance task arithmetic. For instance, \cite{yadav2024ties} improve fusion performance by removing smaller parameters and symbolic conflict parameters during the fusion of task vectors. \cite{yu2024language} employ a random mask to ensure parameter sparsity, thereby reducing noise during fusion. \cite{yang2023adamerging} optimize the coefficients of task vector fusion using test set data to enhance fusion performance. A recent work, PCB-Merging \cite{du2024parameter}, proposes making trade-offs in task fusion by measuring competition between parameters. However, the following issues are generally overlooked in these works:

\noindent 
\textbf{Diversity in Parameter Importance:} Different tasks have unique dependencies on model parameters; parameters crucial for one task may be less significant or irrelevant for another. Many existing methods overlook the potential improvements that accurate parameter importance measures can bring to task arithmetic, resulting in suboptimal performance.

\noindent 
\textbf{Over-reliance on Hyperparameter Tuning:} The aforementioned improvements based on task arithmetic still rely on coefficients in task vector fusion. Although these works propose methods for automatic optimization of these coefficients, concerns about generalization due to this dependency remain unresolved.

\noindent 
\textbf{Other Abilities of Task Arithmetic are Ignored:} While there are numerous studies based on task arithmetic, to the best of our knowledge, only \cite{ilharco2022editing} consider task forgetting by directly subtracting a task-specific task vector from the parameters. However, some noisy elements can significantly impair the overall model function after the subtraction, leading to suboptimal results. This limitation greatly restricts the powerful potential of task arithmetic itself.

To address these challenges, we introduce \textbf{\underline{S}}elective \textbf{\underline{T}}ask \textbf{\underline{A}}rithmetic \underline{\textbf{(STA)}}, a novel training-free framework designed to enable efficient parameter fusion in multi-task learning. STA combines a loss-sensitive parameter importance metric, derived from a first-order Taylor expansion, with a gating mechanism for task-specific parameter selection (see \autoref{fig:intro}). Additionally, we applied this technique to the task forgetting scenario and obtained promising results. The main contributions of STA are: \textbf{\underline{(i)}} a loss-sensitive metric based on first-order Taylor expansion, which provides a precise and computationally efficient assessment of parameter importance for each task; \textbf{\underline{(ii)}} leveraging precise parameter importance measurement, STA can accurately sparsify the task vector of each downstream task model, making it fully suitable for arbitrary task arithmetic operations. This high sparsity also eliminates dependency on the choice of coefficients when performing task arithmetic operations; and \textbf{\underline{(iii)}} our STA method surpasses the current state-of-the-art in different arithmetic tasks, including task fusion and task forgetting.

\begin{figure*}[t]
  \centering
  \includegraphics[width=0.9\linewidth]{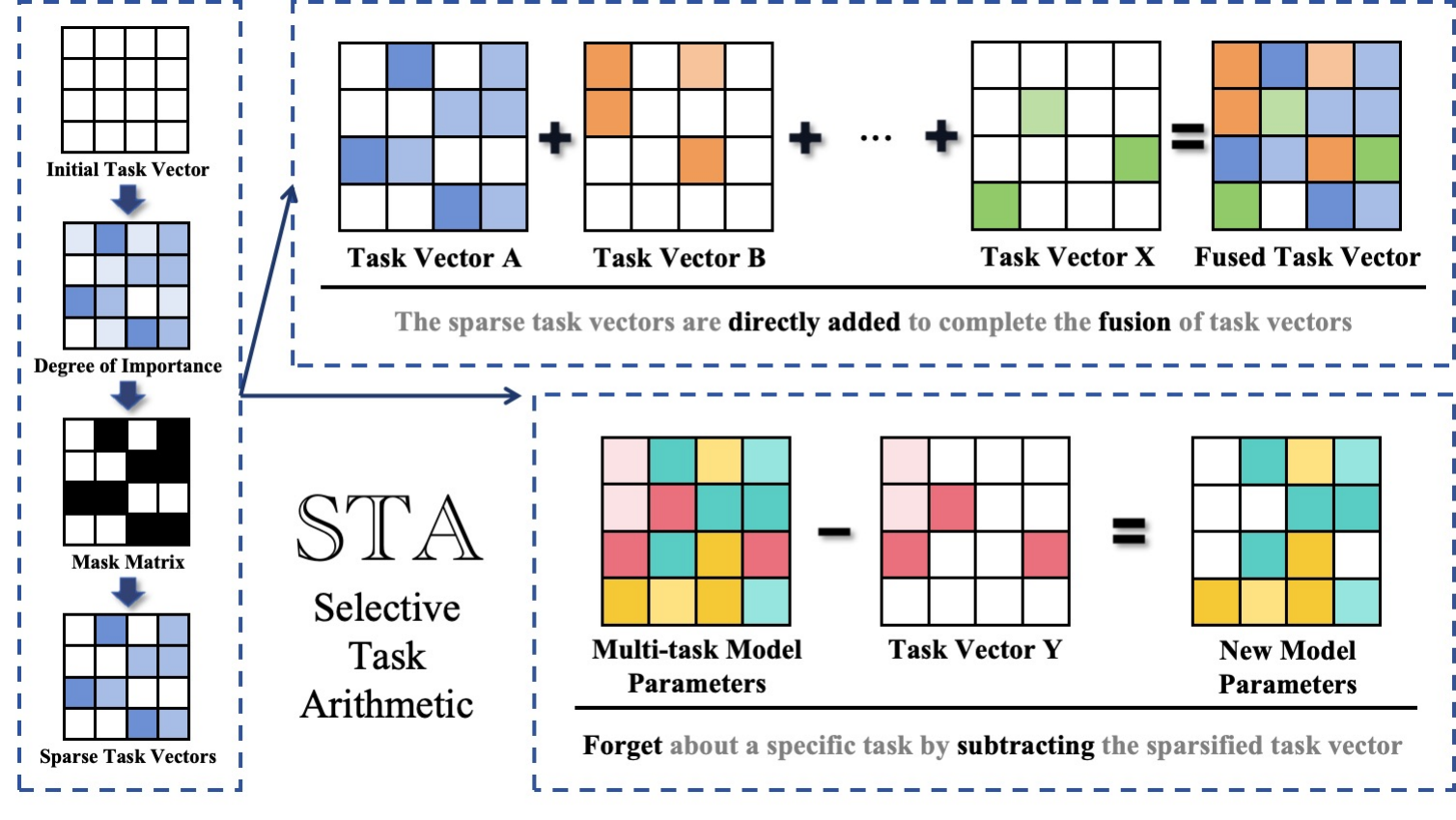}
  \caption{The overview of our proposed Selective Task Arithmetic. In STA, we measure the importance of the task vector through the loss-sensitive parameter importance measurement method, complete the sparsification of the task vector through the importance to obtain the sparse task vector, and then use this task vector to complete tasks such as task fusion or specific task forgetting. A more detailed description of the process is provided in \autoref{sec:method}.}
  \label{fig:overview}
\end{figure*}

%% file: TaskVectorX/sec/3_prelim.tex
\section{Preliminary}
\label{sec:preliminary}

To better contextualize our work, we first introduce the target problem of our study in Section \ref{sec:taskdef}, followed by a discussion of the commonly used solutions in the current literature.

\subsection{Task Definition}
\label{sec:taskdef}
Our objective is to integrate multiple models fine-tuned on different downstream tasks that share identical pretrained weights, without requiring further training. We aim to maintain the original model architecture without adding new parameters, thus creating a unified model capable of executing multiple downstream tasks concurrently. Additionally, we explored the ability to remove a task from a model, a process we refer to as task forgetting. These tasks can be formalized as follows.

\noindent \textbf{Fine-tuning} Consider a scenario where we have pretrained weights for a given architecture $\Theta_{pre} \in \mathcal{A}$, with $\mathcal{A}$ representing a specific model architecture and $\Theta_{pre}$ denoting the pretrained model parameters. Furthermore, let there be $N$ distinct downstream datasets, each denoted as $\mathcal{D}_i = \{ (x_j, y_j) \mid j \in [1, \mathcal{N}_i] \}$ for $i \in [1, N]$, where $\mathcal{N}_i$ signifies the total number of samples in the $i$-th dataset.

The fine-tuning process on these datasets is mathematically represented as:
\begin{equation}
\mathop{\arg\min}\limits_{\Theta_i} \frac{1}{\mathcal{N}_i} \sum_{j=1}^{\mathcal{N}_i} \mathcal{L}(x_j, y_j | \Theta_i), \quad i \in [1, N]
\end{equation}
where $\mathcal{L}(\cdot)$ denotes the loss function employed during fine-tuning. This process yields $N$ models, $\Theta_1, \ldots, \Theta_N$, each fine-tuned on its respective downstream task.

\noindent \textbf{Training-Free Task Fusion:} Upon obtaining the fine-tuned models, our main goal is to merge the parameters of all downstream tasks into a single model. This fusion process can be formulated as:
\begin{equation} 
\Theta_{fused} = \mathcal{M}([\Theta_{pre}, \Theta_1, \ldots, \Theta_N]) 
\end{equation} 
where $\mathcal{M}$ denotes the algorithmic process of merging the models into one. All models, both before and after fusion, adhere to the same architecture; only the parameters vary:
\begin{equation} 
\Theta_{pre}, \Theta_1, \ldots, \Theta_N, \Theta_{fused} \in \mathcal{A}
\end{equation} 

We anticipate that the fused model $\Theta_{fused}$ will retain the capability to effectively handle all downstream tasks. This objective can be formally articulated as:
\begin{equation}
\mathop{\arg\min}\limits_{\mathcal{M}([\Theta_1, \ldots, \Theta_N])} \sum_{i=1}^{N} \sum_{j=1}^{\mathcal{N}i} \mathcal{L}(x_j, y_j | \Theta_{fused}) 
\end{equation} 

In essence, our central challenge is to devise a model fusion algorithm, denoted as MM, capable of integrating the parameters of multiple models while ensuring they retain their ability to perform their respective downstream tasks.

\noindent \textbf{Training-Free Task Forgetting:} In contrast to most prior work that exclusively focuses on parameter-level model fusion, we introduce the concept of parameter-level task forgetting, as initially proposed in \cite{ilharco2022editing}.  This approach has not received much attention in subsequent research \cite{du2024parameter, yang2023adamerging, yu2024language, yadav2024ties}. The forgetting task can be defined as follows: given a downstream model with parameters $\Theta_i$ and a pretrained model with parameters $\Theta_{pre}$, we define a computational process $\mathcal{F}(\cdot)$ to generate a "forgotten" model, $\Theta_{neg}$:
\begin{equation} 
\Theta_{neg} = \mathcal{F}(\Theta_i, \Theta_{pre}) 
\end{equation} 

The goal is for $\Theta_{neg}$ to preserve performance on tasks unrelated to task $i$ while substantially reducing accuracy on task $i$. This can be formally stated as:
\begin{equation}
\mathop{\arg\min}\limits_{\mathcal{F}(\Theta_i, \Theta_{pre})} \sum_{m=1}^{\mathcal{N}j} \mathcal{L}(x_m, y_m | \Theta_{neg}) - \sum_{n=1}^{\mathcal{N}i} \mathcal{L}(x_n, y_n | \Theta_{neg}) 
\end{equation} 
where $i \neq j$, meaning that tasks $i$ and $j$ are distinct.

\subsection{Task Arithmetic} 

Drawing inspiration from \cite{frankle2020linear, wortsman2022robust, matena2022merging, wortsman2022model, li2022branch, ainsworth2022git, don2022cold}, the authors observed that when multiple models share the same optimization path (i.e., the same pretraining weights), a practical approach for editing model parameters is through linear interpolation. Following this, \cite{ilharco2022editing} introduced a method known as "task arithmetic" to adjust model parameters for multi-task fusion and task-specific forgetting.

The core idea behind task arithmetic involves calculating the difference between the fine-tuned model parameters and the pretrained model parameters, referred to as the task vector: 
\begin{equation} 
\label{equ} 
\delta_i = \Theta_i - \Theta_{pre}, \quad i \in [1, N] 
\end{equation} 
Here, $\delta_i$ represents the task vector corresponding to the $i$-th downstream task.

In their method, task vectors are utilized to flexibly perform both multi-task fusion and task-specific forgetting. Specifically, the sum of these task vectors is added to the pretrained weights to obtain the fused model: \begin{equation} 
\label{equ:taskvector}
\Theta_{fused} = \Theta_{pre} + \lambda \sum_{i=1}^N \delta_i 
\end{equation} 
where $\lambda$ is a hyperparameter that modulates the sum of the task vectors.

Similarly, task forgetting is achieved by subtracting a task vector from the pretrained model: 
\begin{equation} 
\Theta_{neg} = \Theta_{pre} - \gamma \cdot \delta_i 
\end{equation} 
Here, $\gamma$ is a hyperparameter that controls the degree of forgetting.

While this approach offers an intuitive solution to the problems outlined in Section \ref{sec:taskdef}, it has inherent limitations, as it overlooks the error introduced by directly adding or subtracting task vectors.


%% file: TaskVectorX/sec/3_method.tex
\section{STA: Our Method}
\label{sec:method}

Our fine-tuning process aligns with the method detailed in \autoref{sec:preliminary} and is therefore not reiterated here. Let us assume the existence of $N$ downstream models corresponding to $N$ downstream tasks, denoted as $\Theta_1, \ldots, \Theta_N$. Furthermore, suppose we have acquired the task vectors for each downstream task, represented as $\delta_1, \ldots, \delta_N$, as per \autoref{equ:taskvector}. We will proceed by outlining our approach step by step.

\subsection{Measure the Importance of Model Parameters}
Previous research \cite{zhong2024convolution, yinjunk, wang2024model, zhong2023seeking, lubana2020gradient} has identified significant redundancy in the parameters of contemporary deep learning models. These redundant parameters can introduce substantial noise during model fusion, potentially hindering the performance of the fused model. Minimizing this noise could significantly enhance the fusion model's performance.

Drawing inspiration from this perspective, we initially assess the importance of model parameters prior to the fusion process. After thorough evaluation, we primarily adopt an importance measurement on model parameters based on the model performance preservation in downstream tasks. Specifically, the importance of a parameter in the model is quantified by the following formula:
\begin{equation}
    I_{i,j} =|\theta_i^\top \nabla_{\theta_i}\mathcal{L}(x_j,y_j|\Theta)|
\end{equation}
where $I_{i,j}$ represents the importance of the $i$th parameter $\theta_i$ relative to sample $j$ and the absolute value is taken for the purpose of measuring the amplitude. This formula can be deduced from the following equation:
\begin{equation}
    \mathcal{L}(x_j,y_j|\Theta) - \mathcal{L}(x_j,y_j|\Theta-\theta_i) \approx \theta_i^\top \nabla_{\theta_i}\mathcal{L}(x_j,y_j|\Theta)
\end{equation}
where $\theta_i$ is a component of $\Theta$. This result is derived from the first-order Taylor expansion of the function on the left indicating the change value of the loss function on sample $j$ when the component $\theta_i$ is removed from the model parameter $\Theta$.

If we have a small sample set $\mathcal{S} = \{(x_j,y_j)|j\in[1,N_\mathcal{S}]\}$, then we can compute the cumulative importance on that sample set as defined above for $\theta_i$:
\begin{equation}
    I_{i,\mathcal{S}} = \sum_{j=1}^{N_\mathcal{S}} \mathop{abs}(\theta_i^\top \nabla_{\theta_i}\mathcal{L}(x_j,y_j|\Theta))
\end{equation}

According to this definition, we can calculate the importance of each parameter of the $N$ downstream models, where the samples required for the calculation can be extracted from the training dataset of each task itself (see \autoref{sec:exp} for details).

Now, assuming that we have calculated the importance of each parameter of each downstream task model through the above method, we can record the parameter importance matrix of each model $I_i, i\in[1,N]$. Since each parameter in the model has its corresponding level of importance, $I$ and $\Theta$ should belong to the same architecture ($I_i,\Theta_i \in \mathcal{A}$).

\subsection{Crossmasking}
Since the task vector is only obtained by subtracting the pretrained weight from the weight of the fine-tuned model, we can assume that the importance of the fine-tuned model parameters can also represent the importance of the corresponding task vector at that bit.

A hunch is that \textbf{we can discard the relatively unimportant parts of each task vector and keep only the important parts}, so as to minimize the errors introduced by fusion.

However, this also brings a new problem, the importance measures on different downstream task models are not limited to the same value range, cannot be directly compared with each other, and naturally cannot make fair trade-offs.

To solve this problem, we normalize the importance of each parameter according to the hierarchical structure of the model:
\begin{equation}
    I^l = \frac{ I^l - \mathop{\min}(I^l)}{\mathop{\max}(I^l) - \mathop{\min}(I^l)}
\end{equation}
where $I^l$ represents the importance matrix of the current layer $l$ parameters.

When the above importance matrix is normalized within itself, we can use the normalized importance value to select the task vector $\delta_1,\ldots,\delta_N$, the specific selection strategy can be expressed as:
\begin{equation}
    I^*_{1'},\ldots,I^*_{N'} = \mathop{sort}(I^*_1,\ldots,I^*_N) 
\end{equation}
where $I^*_1,\ldots,I^*_N$ represents the importance value of the element at the same location as these task vectors, The purpose of this step is to take out the importance values at the same position of $N$ task vectors and sort them from smallest to largest to form a new sequence $I^*_{1'},\ldots,I^*_{N'}$, which is applicable to any element.

Next, we can calculate the quantile value of this $N$ importance values in this position:
\begin{align}
    L_p &= 1 + (N-1)\cdot p \\
    I^*_{p} &= I^*_{\mathop{int}(L_p)'} + (I^*_{(\mathop{int}(L_p) +1)'} - I^*_{\mathop{int}(L_p)'}) \cdot (L_p - \mathop{int}(L_p))
\end{align}
where $p$ is a hyperparameter, and $p\in [0,1]$. When the value of $p$ increases, the corresponding quantile value $I^*_p$ also increases, that is, it tends to be the maximum value of the current position.

The quantile values generated by each bit of the importance matrix $I$ according to $p$ can form a threshold matrix $\mathcal{T}$ with the same structure as $I$, can be expressed as $\mathcal{T},I \in \mathcal{A}$. This threshold matrix can be used to generate a mask matrix, which is represented below:
\begin{equation}
    M_i^* = \mathbb{I}(I^*_i > \mathcal{T}^*), i\in[1,N]
\end{equation}
where $*$ still means any position, $\mathbb{I}(\cdot)$ represents the indicator function, which has a value of 1 when the conditions in it are met and a value of 0 when the conditions in it are not met.

After generating the corresponding mask values for each element at each position according to the above method, we get $N$ mask matrices $M_1,\ldots,M_N$. We can directly multiply the mask matrix and its corresponding task vectors bitwise to filter out the values that are not important enough in the task vectors:
\begin{equation}
    \delta_{i'} = M_i \odot \delta_i,i\in[1,N]
\end{equation}

After this step, we have processed the original $\delta_1,\ldots,\delta_N$ as $\delta_{1'},\ldots,\delta_{N'}$ through the importance filtering step, and then we can rely on B to solve the problem raised in Section \ref{sec:preliminary}.

\subsection{Task fusion}
We complete model fusion by adding the filtered task vectors and applying them on the pretrained parameters, which can be expressed as:
\begin{equation}
    \Theta_{fused} = \Theta_{pre} + \sum_{i=1}^N \delta_{i'}
\end{equation}
\textbf{It's worth noting that we have one less hyperparameter $\lambda$ than \autoref{equ:taskvector}}, because our importance-based masking approach results in very high sparsity for different task vectors, so we don't need to adjust the amplitude via $\lambda$.

\subsection{Task Forgetting}
Similar to the fusion of models, we complete the forgetting of a specific task by subtracting the filtered task vector:
\begin{equation}
    \Theta_{neg} = \Theta_{k} - \delta_{i'}
\end{equation}
The above equation completes the operation of forgetting task $i$ from model $k$, and, as before, it does not require a $\gamma$ to control the amplitude.

The sole distinction with task fusion is that, when calculating quantile values, task fusion considers the importance of parameters from different models at the current position, whereas task forgetting calculates the quantile value based on the importance of its own parameters in the current layer. Additionally, it determines which parameter values need to be excluded in the current layer based on the quantile value.

\begin{table*}[h]
\centering
\resizebox{\linewidth}{!}{
\begin{tabular}{cccccc}
\hline
Method          & \begin{tabular}[c]{@{}c@{}}Core formula\end{tabular} & \begin{tabular}[c]{@{}c@{}}Considering \\ task forgetting\end{tabular} & \begin{tabular}[c]{@{}c@{}}Considering \\ model fusion\end{tabular} & \begin{tabular}[c]{@{}c@{}}Depends on \\ scaling coefficient\end{tabular} & \begin{tabular}[c]{@{}c@{}}Rely on \\ test set data\end{tabular} \\ \hline
\rowcolor[HTML]{EFEFEF} 
Task Arithmetic\cite{ilharco2022editing} & $\delta_i = \Theta_i - \Theta_{pre}, \quad i \in [1, N]$                                                       & {\color[HTML]{32CB00} Yes}                                             & {\color[HTML]{32CB00} Yes}                                          & {\color[HTML]{32CB00} Yes}                                                & {\color[HTML]{CB0000} No}                                        \\
\rowcolor[HTML]{EFEFEF} 
TIES-Merging\cite{yadav2024ties}    & -                                                       & {\color[HTML]{CB0000} No}                                              & {\color[HTML]{32CB00} Yes}                                          & {\color[HTML]{32CB00} Yes}                                                & {\color[HTML]{CB0000} No}                                        \\
\rowcolor[HTML]{EFEFEF} 
AdaMerging\cite{yang2023adamerging}      & $\mathop{\min}\limits_{\lambda_1,\lambda_2,\ldots,\lambda_K}\sum_{k=1}^K\sum_{x_i\in\mathcal{B}_k}H(f_{\theta_{MTL}}(x_i))$                                                       & {\color[HTML]{CB0000} No}                                              & {\color[HTML]{32CB00} Yes}                                          & {\color[HTML]{32CB00} Yes}                                                & {\color[HTML]{32CB00} Yes}                                       \\
\rowcolor[HTML]{EFEFEF} 
DARE\cite{yu2024language}            & $\widetilde{\delta}^{t}=\left(1-m^{t}\right) \odot \delta^{t}$                                                       & {\color[HTML]{CB0000} No}                                              & {\color[HTML]{32CB00} Yes}                                          & {\color[HTML]{32CB00} Yes}                                                & {\color[HTML]{CB0000} No}                                        \\
\rowcolor[HTML]{EFEFEF} 
PCB-Merging\cite{du2024parameter}             & $\beta_{\text {intra }, i}=\operatorname{Softmax}\left(N * \operatorname{Norm}\left(\tau_{i} \odot \tau_{i}\right)\right)$                                                       & {\color[HTML]{CB0000} No}                                              & {\color[HTML]{32CB00} Yes}                                          & {\color[HTML]{32CB00} Yes}                                                & {\color[HTML]{CB0000} No}                                        \\
\rowcolor[HTML]{EFEFEF} 
SPA(\textbf{Ours})       & $I_{i,j} = \mathop{abs}(\theta_i^\top \nabla_{\theta_i}\mathcal{L}(x_j,y_j|\Theta))$                                                       & {\color[HTML]{32CB00} Yes}                                             & {\color[HTML]{32CB00} Yes}                                          & {\color[HTML]{CB0000} No}                                                 & {\color[HTML]{CB0000} No}                                        \\ \hline
\end{tabular}
}
\caption{Comparison with different baselines}
\label{tab:feature}
\end{table*}

\subsection{Extension}
For the measure of importance, inspired by \cite{lubana2020gradient}, we set up two other measures of importance,

\noindent \textbf{Based on Amplitude} Since the small amplitude value parameter has little influence on the activation value of the model, similar to many model pruning work, we explore a method of measuring importance from the perspective of amplitude value, which can be formally expressed as:
\begin{equation}
    I_i = \parallel \theta_i \parallel^2_2
\end{equation}
where $I_i$ denotes the importance of the $i$th parameter $\theta_i$.

\noindent \textbf{Mixed Approach} In order to balance the two measurement methods of loss retention and parameter size, we also tried to balance the two methods, which is called a mixed approach. This metric can be expressed as follows:
\begin{equation}
    I_{i,j} = \mathop{abs}(\theta_i^\top \nabla_{\theta_i}\mathcal{L}(x_j,y_j|\Theta)) \cdot \parallel \theta_i \parallel^2_2
\end{equation}
we expect this measure to combine the characteristics of both amplitude and loss retention metrics to achieve a more accurate measure of importance.

%% file: TaskVectorX/sec/4_exp.tex
\section{Task Fusion Experiments}
\label{sec:exp}

\subsection{Setup}
\label{sec:expset}

\noindent \textbf{Datasets} To verify the versatility of our method, we completed our experiments on six common computer vision datasets, namely MNIST\cite{deng2012mnist}, GTSRB\cite{Stallkamp2012}, Cars\cite{6755945}, SVHN\cite{Netzer2011ReadingDI}, DTD\cite{cimpoi14describing}, EuroSAT\cite{helber2019eurosat,helber2018introducing}. 

\noindent \textbf{Baselines} For the selection of baseline, we only consider the traditional model parameter averaging method, the task arithmetic method, and the improved method for task arithmetic. These include TIES-Merging\cite{yadav2024ties}, which performs patterned noise cancellation when fusing multiple task vectors, DARE\cite{yu2024language} methods, which uses random masks to circumvent fusion noise, and PCB\cite{du2024parameter} method, which take advantage of the competition between parameters to optimize fusion results. DARE can be used as a pre-processing method for any of these methods, and we will take these into account in subsequent experiments. At the same time, because the AdaMerging\cite{yang2023adamerging} method requires access to the test set data, which is difficult to meet under normal circumstances, we did not include this method in the comparison experiments. We have sorted out all the characteristics of the baseline mentioned above and the core formulas to better reflect the differences between our approach and other baseline methods, as shown in \autoref{tab:feature}.

\noindent \textbf{Fine-tuning Settings} The settings on our fine-tuning were consistent with the settings in \cite{ilharco2022editing}. We trained ViT-B/16\cite{dosovitskiy2020image} on a single NVIDIA A800 at a learning rate of 1e-5, 5 rounds for the MNIST dataset, 35 rounds for Cars, 77 rounds for DTD, 12 rounds for EuroSAT, 11 rounds for GTSRB, and 4 rounds for SVHN.

\noindent \textbf{Hyperparameter Settings} For the task arithmetic method, we only set the coefficient at the time of their fusion to the recommended value in their paper(ie. $\lambda = 0.4$). For the TIES-Merging method, we set the mask ratio for the small amplitude parameter value to $0.8$, and the other settings are consistent with the task arithmetic. As for the PCB method, we use the default parameter settings of their official code. For our STA method, we uniformly set the quantile hyperparameter $p=0.7$ regardless of the number of task fusions, and we always keep the fusion coefficient at $1$, and do not optimize this coefficient to ensure greater versatility and avoid over-tuning. When using the loss protection metric, the STA method needs to take a small part of the training samples for each task to calculate the loss, and we randomly take 32 samples from the training set to calculate the importance degree for any task.

\subsection{Dual-task Fusion}

\begin{figure*}[t]
  \centering
  \includegraphics[width=\linewidth]{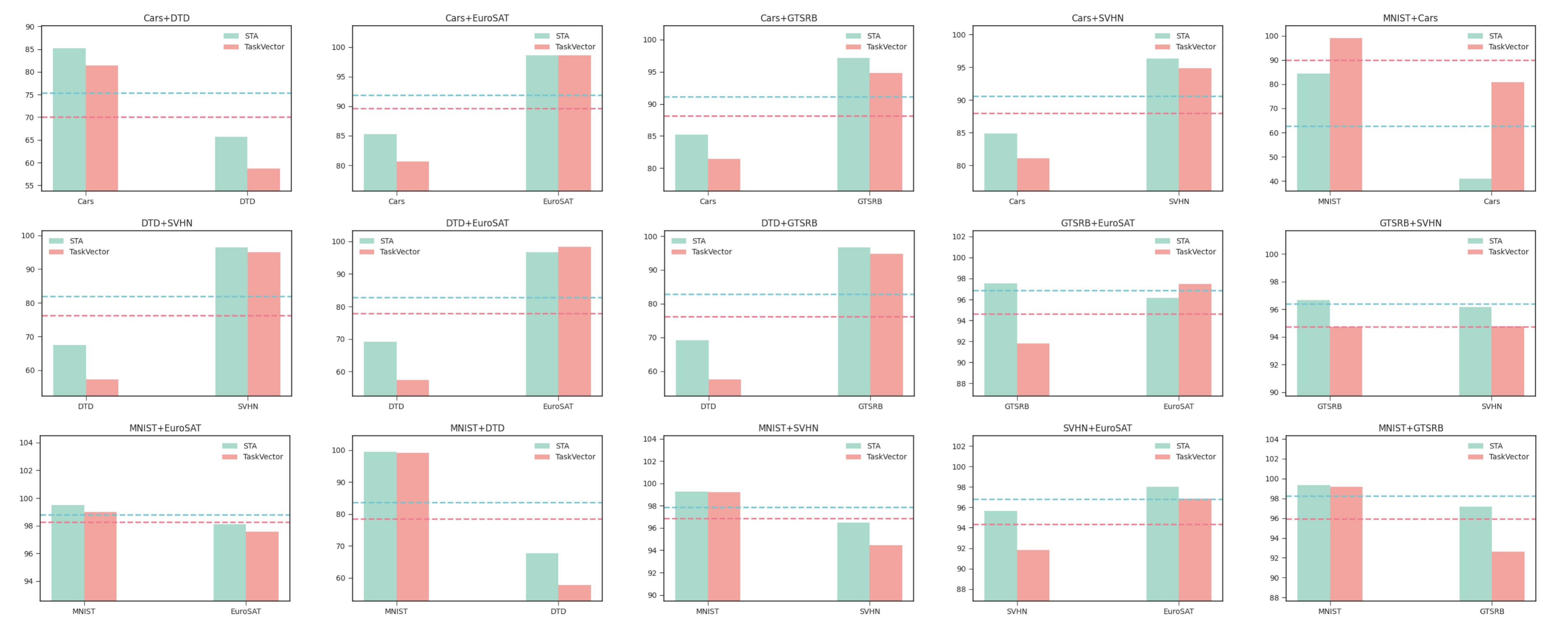}
  \caption{This is a comparison chart of the effect of the two tasks in the case of fusing the two tasks with the task arithmetic method, where the vertical axis represents the accuracy of the test set}
  \label{fig:dual-task}
\end{figure*}

In order to preliminarily evaluate the effectiveness of our method, we combined the six datasets described above in pairs to form a dual-task merging scenario and compared the performance of our method with that of the task arithmetic method, as shown in \autoref{fig:dual-task}. The light green dotted line represents the average accuracy of the proposed STA method on the two tasks, while the pink dotted line indicates the average accuracy of the task arithmetic method on the same tasks. It can be observed that in almost all task combinations, our method significantly outperforms the task arithmetic method, effectively demonstrating the efficacy of our approach.

\subsection{Multi-task Fusion}

\begin{table*}[h]
\centering
\begin{tabular}{cccccccc}
\toprule[0.5mm]
                & MNIST & Cars  & SVHN  & GTSRB & DTD   & EuroSAT & Avg   \\
\midrule
Avg \scriptsize{(ICML2022\cite{wortsman2022model})}             & 96.61 & 71.52 & 79.64 & 67.96 & 46.08 & 77.37   & 73.19 \\
Task Arithmetic \scriptsize{(ICLR2023\cite{ilharco2022editing})} & 98.97 & 67.19 & \underline{91.60}  & 80.71 & 48.34 & 64.59   & 75.23 \\
TIES-Merging \scriptsize{(NIPS2023\cite{yadav2024ties})}            & 53.92 & 75.66 & 76.16 & 76.61 & 49.73 & 90.07   & 70.35 \\
\midrule
Avg \scriptsize{(ICML2022\cite{wortsman2022model})}+DARE        & 96.61 & 71.62 & 79.63 & 68.61 & 45.86 & 77.74   & 73.34 \\
Task Arithmetic \scriptsize{(ICLR2023\cite{ilharco2022editing})}+DARE & 9.85 & 70.22 & 31.83 & 74.93 & \textbf{53.81} & \textbf{92.67} & 55.55 \\
TIES-Merging \scriptsize{(NIPS2023\cite{yadav2024ties})}+DARE       & 52.95 & 75.55 & 76.33 & 76.29 & 49.30  & \underline{90.63}   & 70.17 \\
PCB-Merging \scriptsize{(NIPS2024\cite{du2024parameter})}             & 98.89 & 77.78 & 89.31 & \underline{85.05} & 53.06 & 90.30    & \underline{82.39} \\
\midrule
STA [\textbf{Ours}]\scriptsize(Amp)       & 98.32 & 73.21 & 88.88 & 76.72 & 48.23 & 76.26   & 76.93 \\
STA [\textbf{Ours}]\scriptsize(LP+Amp)    & \underline{98.91} & \underline{79.11} & 90.72 & 80.67 & \underline{53.17} & 83.41   & 80.94 \\
STA [\textbf{Ours}]\scriptsize(LP)        & \textbf{99.34} & \textbf{85.45} & \textbf{93.31} & \textbf{86.61} & 51.99 & 80.37   & \textbf{82.84} \\
\bottomrule[0.5mm]
\end{tabular}
\caption{Comparison of the effectiveness of our method with the baseline method in the multi-task fusion scenario, where \textbf{bold} represents the best result and \underline{underline} indicates the suboptimal outcome}
\label{tab:compare}
\end{table*}

We compared the multi-task fusion scenario with a baseline selected in \autoref{sec:expset}. The results were shown in \autoref{tab:compare}, where DARE can be used primarily for preprocessing, so we report the effects of different other methods using DARE versus not using DARE, except for the PCB-Merging method, which explicitly reports in the paper that no DARE is used for preprocessing on visual tasks.

For the STA method we propose, we tested several measures proposed in \autoref{sec:method}, where \textbf{Amp} represents the degree of importance of the parameter by amplitude only, \textbf{LP+Amp} means the degree of importance co-measured by loss preservation and amplitude, and \textbf{LP} means that the degree of importance is measured only by loss preservation.

From the results, it can be seen that the STA method we proposed has achieved competitive results regardless of the importance measurement method, among which the loss protection-based measurement method has achieved the best effect. Maintaining an accuracy of more than 90\% on the MNIST dataset and SVHN dataset, and achieving an accuracy of more than 80\% on the three datasets of EuroSAT, Cars, and GTSRB, while the average accuracy rate on all tasks has reached 82.84\%, above baseline methods under all settings. Most of the suboptimal results also appear in the STA's result set, which illustrates the significance of the route of assisting model fusion by measuring the importance of parameters, and it is expected to be extended in the future with more granular metric methods. It is also obvious that the application of DARE to computer vision-related tasks does not improve the fusion model very well.

\section{Task Forgetting Experiments}
Different from previous work, we also explore the superiority of our proposed STA method for task arithmetic in task-forgetting problems, and detailed experimental details will be introduced in this chapter.

\subsection{Setup}
In this task, most of our settings remain unchanged as detailed in \autoref{sec:expset}. We continue to use the difference between the weights of the fine-tuned model and the pretrained model to derive the task vector for the corresponding task. Since we utilize the CLIP classification head, our pretrained model inherently possesses zero-shot capabilities across various datasets. For instance, after adapting the classification head to MNIST, the pretrained model demonstrates a certain level of classification ability for MNIST. Our aim is to subtract the task vector corresponding to MNIST from the pretrained model's weights to minimize its effectiveness on MNIST, while preserving its performance on the original pretrained task, such as ImageNet \cite{5206848}.

\begin{table}[htp]
\centering
\resizebox{0.6\columnwidth}{!}{
\begin{tabular}{cccc}
\toprule[0.5mm]
                         & Method          & Target $\downarrow$ & Control $\uparrow$ \\
                         \midrule
\multirow{3}{*}{MNIST}   & pretrained     & 51.73  & 68.33   \\
                         & Task Arithmetic & 9.24   & 62.3    \\
                         & STA(LP)         & 9.04($\downarrow \mathbf{0.2}$)   & 64.51($\uparrow \mathbf{2.21}$)   \\
                         \midrule
\multirow{3}{*}{Cars}    & pretrained     & 64.74  & 68.33   \\
                         & Task Arithmetic & 9.85   & 57.1    \\
                         & STA(LP)         & 10.25  & 59.85($\uparrow \mathbf{2.75}$)   \\
                         \midrule
\multirow{3}{*}{SVHN}    & pretrained     & 51.99  & 68.33   \\
                         & Task Arithmetic & 6.36   & 55.82   \\
                         & STA(LP)         & 6.33($\downarrow \mathbf{0.03}$)   & 62.62($\uparrow \mathbf{6.80}$)   \\
                         \midrule
\multirow{3}{*}{GTSRB}   & pretrained     & 43.37  & 68.33   \\
                         & Task Arithmetic & 7.49   & 63.07   \\
                         & STA(LP)         & 7.38($\downarrow \mathbf{0.11}$)   & 64.41($\uparrow \mathbf{1.34}$)   \\
                         \midrule
\multirow{3}{*}{DTD}     & pretrained     & 42     & 68.33   \\
                         & Task Arithmetic & 7.84   & 56.39   \\
                         & STA(LP)         & 7.63($\downarrow \mathbf{0.21}$)   & 58.42($\uparrow \mathbf{2.03}$)   \\
                         \midrule
\multirow{3}{*}{EuroSAT} & pretrained     & 54.63  & 68.33   \\
                         & Task Arithmetic & 8.63   & 54.94   \\
                         & STA(LP)         & 8.56($\downarrow \mathbf{0.07}$)   & 62.07($\uparrow \mathbf{7.13}$)   \\
                         \midrule
\multirow{3}{*}{Avg}     & pretrained     & 51.41  & 68.33   \\
                         & Task Arithmetic & 8.23  & 58.27   \\
                         & STA(LP)         & 8.19($\downarrow \mathbf{0.04}$)  & 61.89($\uparrow \mathbf{3.62}$ )  \\
                         \bottomrule[0.5mm]
\end{tabular}
}
\vspace{8pt}
\caption{Comparison of the performance of the STA method and the task arithmetic method on different datasets on task forgetting}
\label{tab:forget}
\end{table}

\begin{figure*}[ht]
  \centering
  \includegraphics[width=0.9\linewidth]{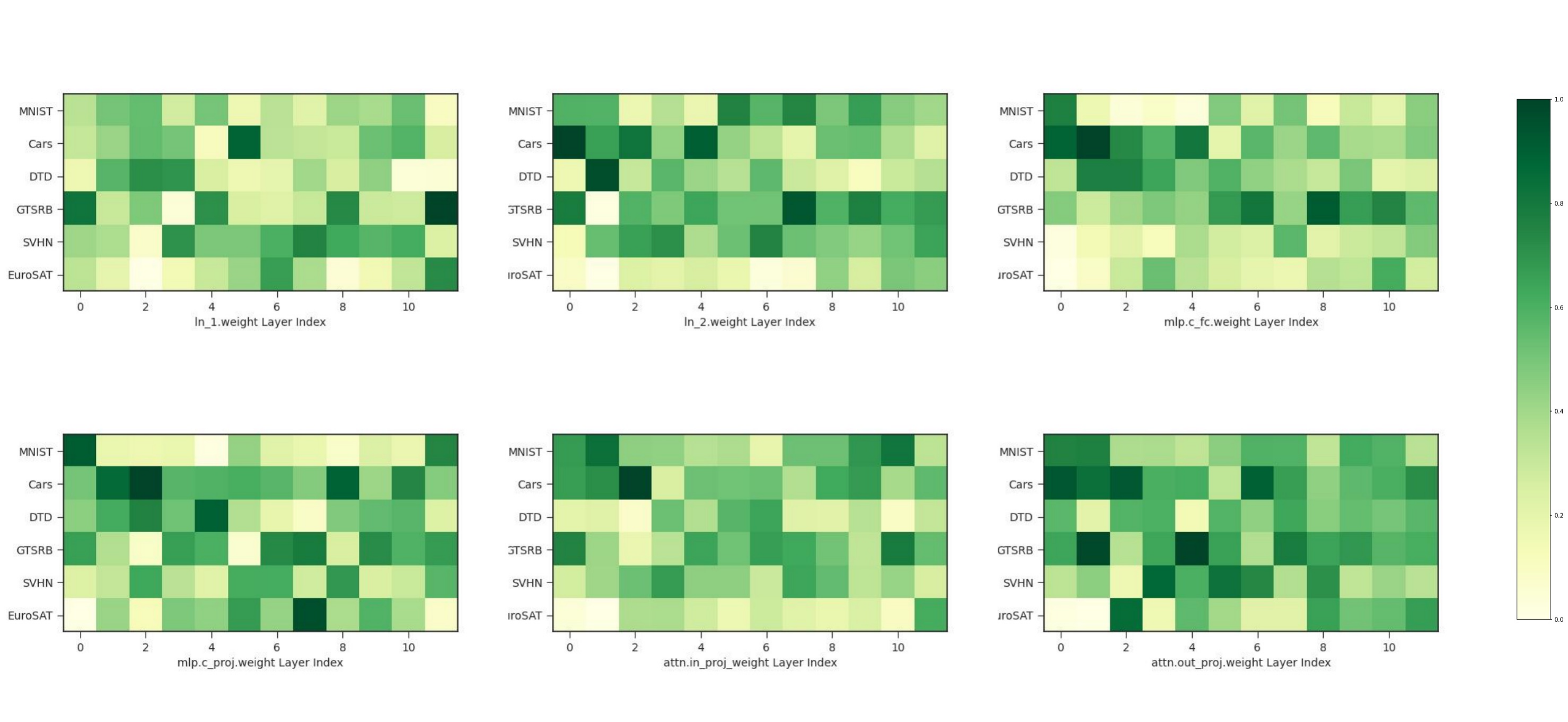}
  \caption{Visualization on different types of model layers for different tasks}
  \label{fig:vis}
\end{figure*}

\subsection{Results}
The results of our experiments are presented in \autoref{tab:forget}. "Target" refers to the task we aim to forget, while "Control" indicates the task we wish to retain during the forgetting process. We evaluated the performance of both the task arithmetic method and our proposed STA method on the Target and Control tasks across six datasets. For reference, we also included the performance of the pretrained model. Finally, we averaged the performance outcomes of the different methods on the Target and Control tasks to provide an overall assessment of the methods.

Our objective is to minimize the performance of the Target task while maximizing the performance of the Control task. The results indicate that our proposed STA method demonstrates superior task forgetting capabilities compared to task arithmetic across nearly all tasks. Specifically, for five out of six tasks, the STA method achieves lower performance on the Target task, indicating more precise forgetting. Concurrently, on the Control task, our method significantly outperforms task arithmetic, offering greater protection for irrelevant tasks. For the remaining task, the performance difference between the STA method and task arithmetic on the Target task is only 0.4, while our method still maintains a substantial lead of 2.75 on the Control task. The final average statistics further illustrate that our method consistently maintains lower performance on the Target task compared to task arithmetic and achieves a performance improvement of 3.62 on the Control task

This outcome can be further elucidated by considering the noise level. Our STA method effectively eliminates less critical parameters in each model based on their degree of importance. Moreover, it substantially reduces the risk of deleting parameters that are relevant to the target task, thereby minimizing harm to the control task. Consequently, the STA method exhibits a significant advantage over the straightforward application of task arithmetic methods.

\subsection{An Intuitive Understanding of STA}

In order to further deepen the understanding of the actual working state of STA, we analyze and visualize the sparsity of each layer of STA after masking under the default hyperparameter setting, which is used to determine the contribution of each task vector to the fusion model in a certain layer.

In order to more clearly see the parameter selection rate of task vectors in different layers after using STA, we divide the parameters of ViT into several different types, namely \textit{ln\_1.weight}, \textit{ln\_2.weight}, \textit{mlp.c\_fc.weight}, \textit{mlp.c\_proj.weight}. The results of the visualization are shown in \autoref{fig:vis}.

As can be seen from the figure, for each type of parameter, the selection rate of STA in each layer is relatively average, and the parameters of each task have a relatively high selection rate in some specific layers, and also have a low selection rate in some layers. This further supports the rationality of our argument that by \textit{selecting important parameters and removing unimportant parameters, we can reduce interference and improve the effectiveness in task arithmetic}.

%% file: TaskVectorX/sec/5_con.tex
\section{Conclusion}
\label{sec:con}
In this work, we propose the STA method to enhance the performance of task arithmetic. The STA method identifies important parameters within the model by evaluating their significance. In the context of multi-task fusion, we make trade-offs based on the varying importance of different task models for each parameter, thereby reducing noise in task fusion. For task forgetting, we apply trade-offs according to the importance of the target's parameters, which helps preserve the model's capability to perform other tasks while minimizing the retention of the target task. Our experiments demonstrate that the STA method achieves SOTA performance across different tasks, validating our hypothesis.